\documentclass{article}
 
\usepackage{microtype}
\usepackage{graphicx}
\usepackage{subfigure}
\usepackage{booktabs} %
\usepackage{subcaption}
\PassOptionsToPackage{hyphens}{url}\usepackage{hyperref}

\usepackage[accepted]{icml2024}

\usepackage{amsmath}
\usepackage{amssymb}
\usepackage{mathtools}
\usepackage{amsthm}

\usepackage[capitalize,noabbrev]{cleveref}
\usepackage{xspace}

\theoremstyle{plain}

\theoremstyle{definition}

\theoremstyle{remark}

\usepackage{paralist}
\usepackage{listings}
\usepackage{xcolor}

\definecolor{codegreen}{rgb}{0,0.6,0}
\definecolor{codegray}{rgb}{0.5,0.5,0.5}
\definecolor{codepurple}{rgb}{0.58,0,0.82}
\definecolor{backcolour}{rgb}{0.95,0.95,0.92}

\definecolor{mydarkred}{rgb}{0.8,0.02,0.02}
\definecolor{mydarkorange}{rgb}{0.40,0.2,0.02}
\definecolor{mypurple}{RGB}{111,0,255}
\definecolor{myred}{rgb}{1.0,0.0,0.0}
\definecolor{mygold}{rgb}{0.75,0.6,0.12}
\definecolor{mydarkgray}{rgb}{0.66, 0.66, 0.66}
\definecolor{mygray}{gray}{0.9}

\lstdefinestyle{mystyle}{
    backgroundcolor=\color{backcolour},   
    commentstyle=\color{codegreen},
    keywordstyle=\color{magenta},
    numberstyle=\tiny\color{codegray},
    stringstyle=\color{codepurple},
    basicstyle=\ttfamily\footnotesize,
    breakatwhitespace=false,         
    breaklines=true,                 
    captionpos=b,                    
    keepspaces=true,                 
    numbers=left,                    
    numbersep=5pt,                  
    showspaces=false,                
    showstringspaces=false,
    showtabs=false,                  
    tabsize=2
}
\newcommand{\ignorethis}[1]{}

\makeatletter
\DeclareRobustCommand\onedot{\futurelet\@let@token\@onedot}
\def\@onedot{\ifx\@let@token.\else.\null\fi\xspace}

\makeatother

\makeatletter

\newcommand\footnoteref[1]{\protected@xdef\@thefnmark{\ref{#1}}\@footnotemark}
\makeatother

\newcommand{\fig}{Fig.~}

\newcommand{\eespeedup}{$2.23\times$\xspace}

\newcommand{\selfspeedup}{$7.03\times$\xspace}

\icmltitlerunning{\method{}: Query-Aware Sparsity for Efficient Long-Context LLM Inference}

\def\method{Quest\xspace}

\newcommand{\kvc}{KV cache\xspace}
\newcommand{\estimate}{Criticality estimation\xspace}
\newcommand{\topk}{Top-K filtering\xspace}
\newcommand{\approxattn}{Approximate attention\xspace}
\newcommand{\qaware}{query-aware}

\begin{document}

\twocolumn[
\icmltitle{\method{}: Query-Aware Sparsity for Efficient Long-Context LLM Inference}

\icmlsetsymbol{equal}{*}

\begin{icmlauthorlist}
\icmlauthor{Jiaming Tang}{equal,sjtu,mit}
\icmlauthor{Yilong Zhao}{equal,sjtu,UW}
\icmlauthor{Kan Zhu}{UW}
\icmlauthor{Guangxuan Xiao}{mit}
\icmlauthor{Baris Kasikci}{UW}
\icmlauthor{Song Han}{mit,nv}
\end{icmlauthorlist}

\icmlaffiliation{sjtu}{Shanghai Jiao Tong University}
\icmlaffiliation{mit}{MIT}
\icmlaffiliation{UW}{University of Washington}
\icmlaffiliation{nv}{NVIDIA}

\icmlcorrespondingauthor{Song Han}{songhan@mit.edu}
\icmlcorrespondingauthor{Baris Kasikci}{baris@cs.washington.edu}

\icmlkeywords{Machine Learning, ICML}

\vskip 0.3in
]

\printAffiliationsAndNotice{\icmlEqualContribution} %

\begin{abstract}
As the demand for long-context large language models (LLMs) increases, models with context windows of up to 128K or 1M tokens are becoming increasingly prevalent. However, long-context LLM inference is challenging since the inference speed decreases significantly as the sequence length grows. This slowdown is primarily caused by loading a large KV cache during self-attention. 
Previous works have shown that a small portion of critical tokens will dominate the attention outcomes. However, we observe the criticality of a token highly depends on the query. To this end, we propose \method, a query-aware \kvc{} selection algorithm. \method keeps track of the minimal and maximal Key values in \kvc pages and estimates the criticality of a given page using Query vectors. By only loading the Top-K critical KV cache pages for attention, \method significantly speeds up self-attention without sacrificing accuracy. We show that \method can achieve up to \selfspeedup self-attention speedup, which reduces inference latency by \eespeedup while performing well on tasks with long dependencies with negligible accuracy loss. Code is available at \url{https://github.com/mit-han-lab/Quest}.
\end{abstract}

\section{Introduction}
The rapid evolution of Large Language Models (LLMs) has shaped our daily lives. With the increasing demand for multi-round conversations and long document queries, the maximum context length of LLMs has dramatically grown from 2K to 1M~\cite{liu2024world,peng2023yarn,tworkowski2023focused}. The 128k context length GPT-4 model has already been deployed in large-scale serving, which is equivalent to 300 pages of text~\cite{openaiannouncement}.  

However, processing long-context requests is challenging. Due to the auto-regressive nature of LLMs, generating one token would require reading the entire \kvc{}. For Llama 7B model~\cite{touvron2023llama} with $32$k context length, the \kvc{} can occupy $16$GB of space, which requires at least $11$ ms to read, which contributes to more than $50$\% of the inference latency\footnote{Tested with FP16 FlashInfer implementation on an RTX4090}, limiting the overall throughput.

Despite the increasingly large size of the \kvc{}, previous works have shown that a small portion of the tokens can dominate the accuracy of token generation~\cite{zhang2023h2o, ge2024model}. Therefore, we can dramatically reduce the inference latency by only loading the critical tokens, while still maintaining accuracy. Thus, it is essential to identify critical portions of the \kvc{}.

\begin{figure}
    \centering
     \includegraphics[width=\linewidth]{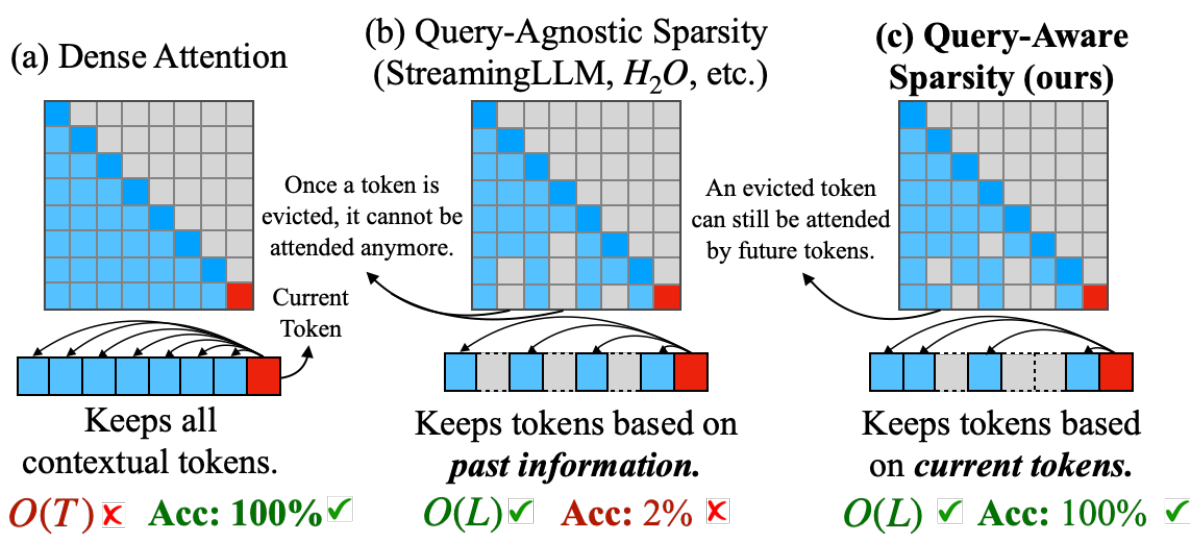}
    \caption{Comparison between Dense Attention(a), Query-Agnostic Sparsity (b) and \method's Query-aware Sparsity (c). \method significantly speeds up self-attention while maintaining high accuracy by dynamically determining the critical tokens based on the current query. $T$ represents the total sequence length and $L$ represents the number of critical tokens for attention.}
    \label{fig:framework}
\end{figure}

In this work, we further observe that the criticality of the tokens can change with different query tokens. As shown in \fig\ref{fig:query}, the critical tokens vary a lot with different queries. Therefore, we need a dynamic and efficient approach to determine which portion of the \kvc{} needs to be attended to. To this end, we propose \method, a query-aware criticality estimation algorithm for long-context LLM inference that efficiently and effectively identifies critical \kvc{} tokens and performs self-attention selectively on chosen tokens, as shown in \fig\ref{fig:framework}.

To reduce the overhead of \kvc{} criticality estimation, \method{} manages \kvc{} at page granularity~\cite{kwon2023efficient}. For each page, \method{} utilizes maximum and minimum values of each feature dimension of the Key vector as the metadata to represent token information. During inference, \method{} considers both the Query vector and the metadata to estimate each page's criticality. Given all criticality scores of the pages, \method{} chooses Top-K pages to perform approximate self-attention, where $K$ is a preset constant (e.g.  128, 256). By reducing the memory movement from the entire \kvc{} to metadata and constant $K$ pages, \method{} significantly accelerates inference.

We evaluate both the accuracy and efficiency of \method{}. Since \method{} dynamically decides the criticality of the tokens, \method{} achieves better accuracy for a given degree of \kvc{} sparsity than baselines on PG19 dataset~\cite{raecompressive2019}, passkey retrieval task~\cite{peng2023yarn}, and LongBench~\cite{bai2023longbench} with $256$ to $4$K token budgets.  For $32$K context, \method{} achieves \selfspeedup self-attention latency reduction compared to FlashInfer~\cite{flashinfer}. Our end-to-end framework demonstrates that \method{} can have \eespeedup inference speedup compared to FlashInfer~\cite{flashinfer} with 4-bit weight quantization. In summary, we make the following contribution:

\vspace{-10pt}
\begin{itemize}
    \setlength{\itemsep}{-3pt}
    \item An analysis of the self-attention mechanism that pinpoints the importance of query-aware sparsity.
    \item \method, an efficient and accurate \kvc{} acceleration algorithm, which exploits query-aware sparsity by dedicated operator designs and implementations.
    \item A comprehensive evaluation of \method, demonstrating up to \selfspeedup self-attention latency reduction and \eespeedup end-to-end latency improvement.
\end{itemize}

\section{Related Work}

\subsection{Long-context Model}

As the demand for long-context models increases, many works have focused on extending the context window of LLMs. Currently, many models utilize Rotary Position Embeddings (RoPE)~\cite{su2023roformer}, and by different scaling methods of RoPE with fine-tuning, the window size of the original 4k Llama-2 has been expanded to 32k for LongChat~\cite{longchat2023} and 128k for Yarn-Llama-2~\cite{peng2023yarn}. Through length extrapolation, the context windows of models reached beyond 1M~\cite{liu2024scaling}.  Beyond open-source models, GPT-4 Turbo supports lengths of up to 128k, while Claude-2 supports up to 200k~\cite{gpt4,claude}. With models increasingly capable of handling long input, this poses challenges for inference efficiency. \method aims to boost long-context inference by exploiting query-aware KV cache sparsity.

\subsection{KV Cache Eviction Algorithm}

For long-context LLM inference and serving scenarios, the huge size of the KV cache results in significant time and space overheads. Many previous efforts have been dedicated to compressing the size of the KV cache to accelerate attention and reduce memory usage. H2O~\cite{zhang2023h2o} retains a limited budget of the important KV cache based on the sum of historical attention scores. FastGen~\cite{ge2024model} further refines the types of tokens, applying a more sophisticated strategy for selecting the KV cache to keep. TOVA~\cite{oren2024transformers} simplifies the policy by deciding which tokens to permanently discard based solely on the current query. StreamingLLM~\cite{xiao2023streamingllm} handles infinitely long texts with attention sinks and a finite KV cache. These methods decide which parts of the KV cache to discard based on historical information or current states, but discarded tokens might be important for future tokens, which may cause the loss of important information. To mitigate this issue, SparQ~\cite{ribar2023sparq} computes approQximate attention scores by channel pruning and selects important tokens through them. However, this approach has not been widely validated for tasks with long dependencies, and the channel-level sparsity might pose challenges to practical acceleration.
Therefore, we propose \method, which retains all of the KV cache and selects part of the KV cache based on the current query to accelerate long-context self-attention without accuracy degradation. 

\section{Methodlogy}
\begin{figure}[t]
    \centering
     \includegraphics[width=0.45\textwidth]{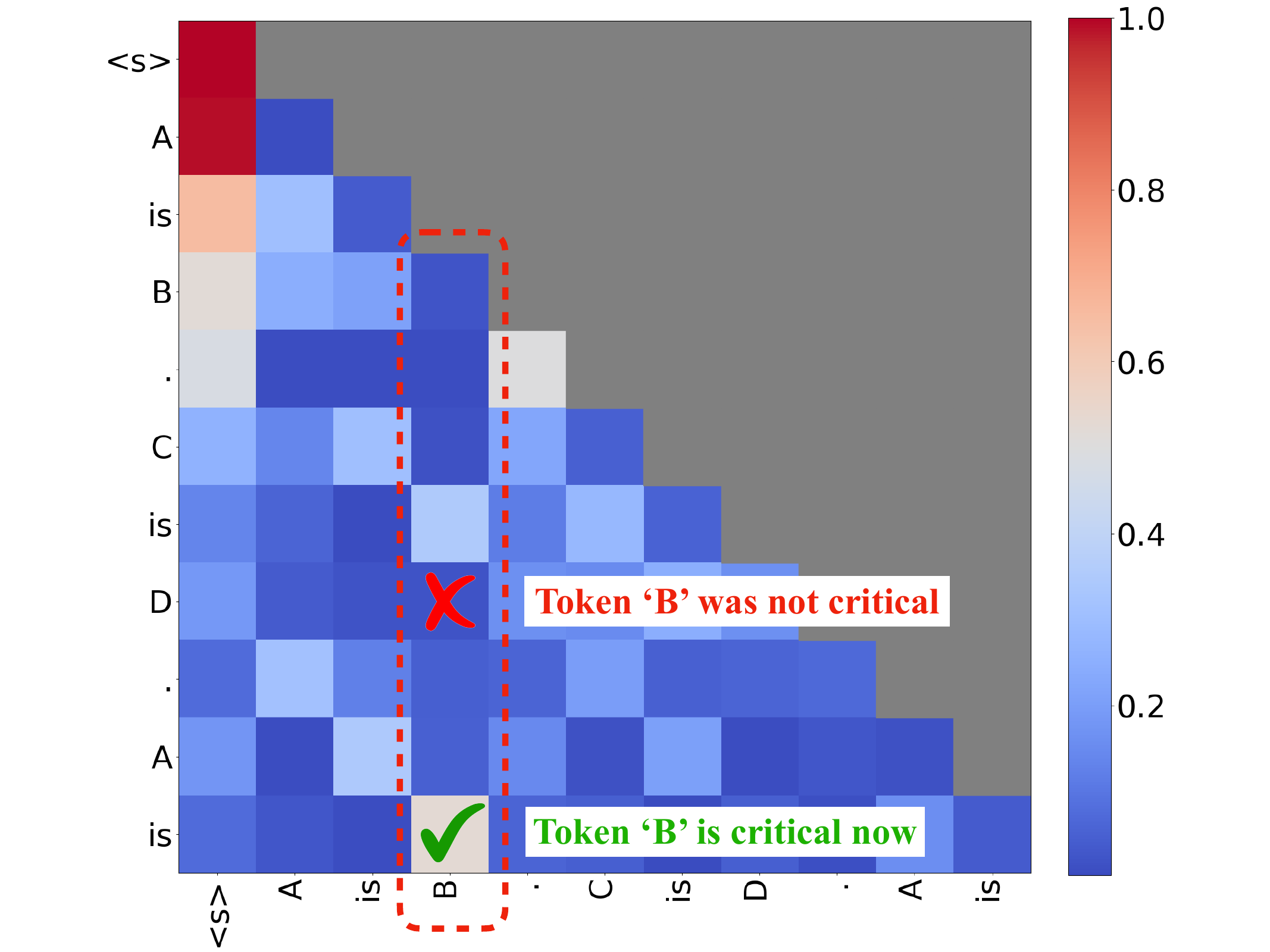}
    \caption{The attention map of prompt ``A is B. C is D. A is''. Each row represents the attention scores of previous tokens queried by the tokens on the left. When queried with ``D'', token ``B'' has a low attention score, showing ``B'' is not critical for generation. However, the ``is'' strongly attends to ``B''. Therefore, the criticality of tokens strongly correlates with the current query token.}
    \label{fig:query}
\vspace{-0.1in}
\end{figure}

In this section, we first motivate \method by analyzing the breakdown of inference cost and self-attention properties. We then present the design of \method and discuss its benefits.

\subsection{Long-context Inference Is Costly}

LLM inference contains two stages, namely, the prefill stage and the decode stage. In the prefill stage, all the input tokens are transformed into embeddings and generate the Key ($K$), Query($Q$), and Value($V$) vectors. Both the Key and the Value vectors are saved in the KV cache for future use. The rest of the prefill stage includes self-attention and feed-forward network (FFN) layers, which produce the first response token. 

\begin{figure}
    \centering
     \includegraphics[width=0.9\columnwidth]{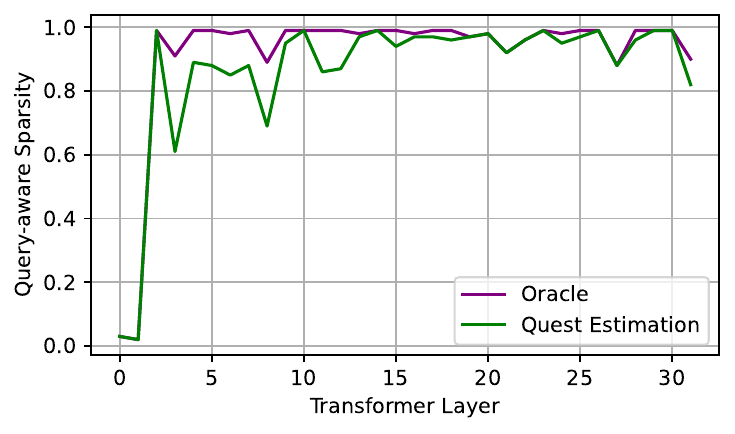}
    \caption{The query aware sparsity for each layer in LongChat-7B model. We measure the sparsity by eliminating KV cache tokens while making sure the perplexity on PG19 increases less than 0.01. For the first two layers, the sparsity is below 10\%, while for the rest of the layers, the sparsity is larger than 90\%, showing great potential for optimization. \method closely aligns with the oracle. }
    \label{fig:qa-sparsity}
\end{figure}

In the decode stage, the model will take the last generated token to calculate its $K,Q,V$. The model uses $Q$ to multiply with every $K$ of previous tokens to generate the \textit{attention weights}. The attention weights will then get normalized using softmax, where each value $a_i$ represents the attention score between $i$th token and the current token. The self-attention layer will output $\sum a_i \cdot V_i$ and send to the FFN.

For one request, the prefill stage only happens once, while a decoding process is needed for every token in the response. Therefore, the decode stage dominates the inference time. For example, for $16$k token prompts and $512$ token responses, over $86$\% of the time is spent on decode stages. Therefore, the decode stage performance is crucial for overall latency.

Moreover, a long-context scenario significantly slows down the decode stage. In every decode stage, the $K$ and $V$ of existing tokens must be loaded to perform self-attention, which can easily reach $16$GB for the $32$k context of Llama-7b\footnote{$\text{KV cache size} = 2 \text{\xspace(both K and V)} * \text{Num of Layer} * \text{Sequence length} * \text{Num of Heads} * \text{Head Dimensions} * \text{Size of FP16} = 2 * 32 * 32 * 32 * 128 * 2 =  16 \text{GB}$}. This memory load operation can take $53$\% of the time in a decode stage. Therefore, optimizing self-attention becomes a must for efficient long-context inference.
\begin{figure}[t]
    \centering
     \includegraphics[width=0.956\linewidth]{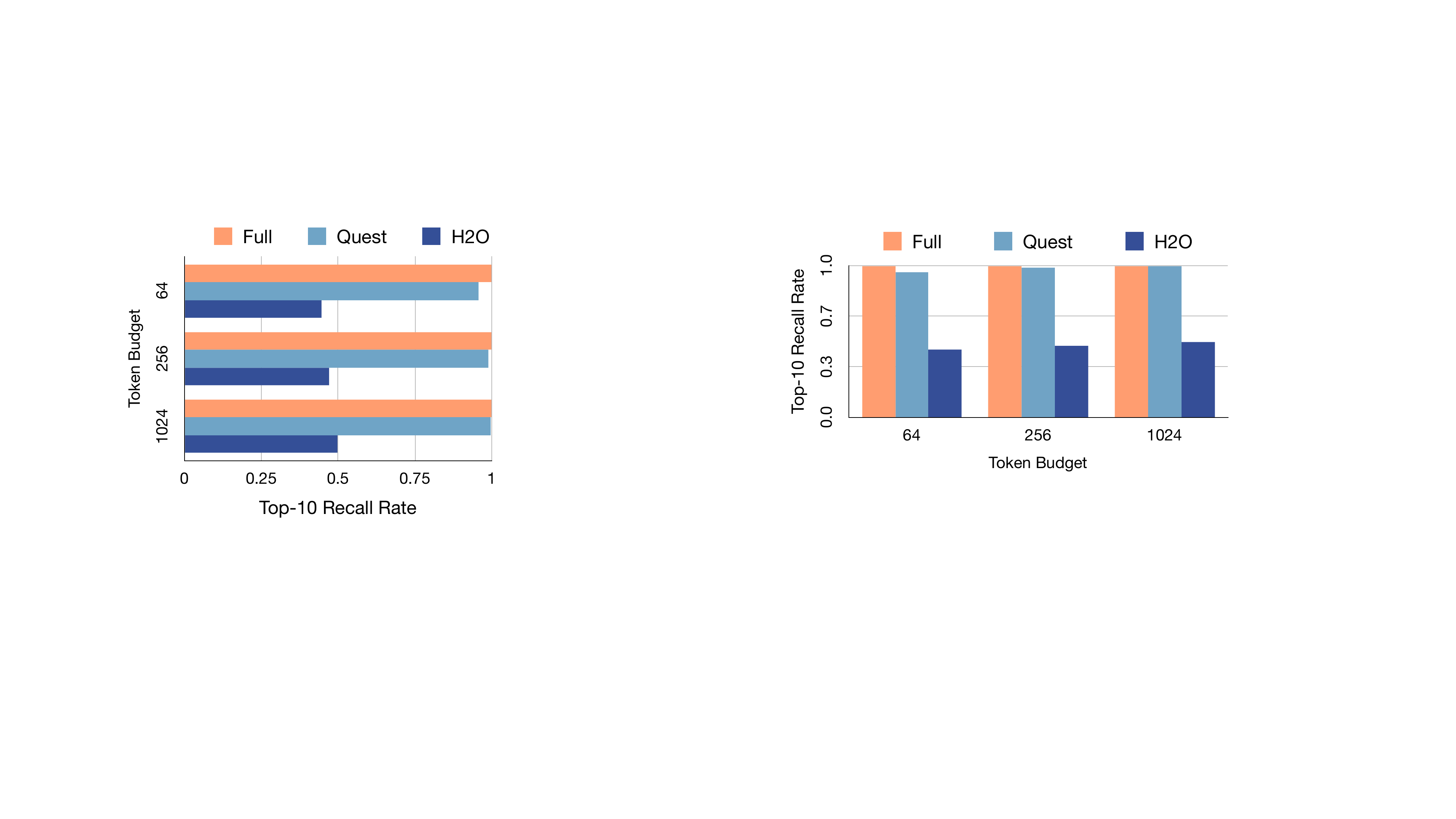}
    \caption{Recall rate of tokens with Top-10 attention scores. Results are profiled with LongChat-7b-v1.5-32k model in passkey retrieval test of $10$K context length. Recall rate is the ratio of tokens selected by different attention methods to tokens selected by the full attention in each round of decoding. The average rate is shown in the figure, with various token budgets assigned.}
    \label{fig:recall}
    \vspace{-0.08in}
\end{figure}

\begin{figure*}[h]
    \centering
     \includegraphics[width=\linewidth]{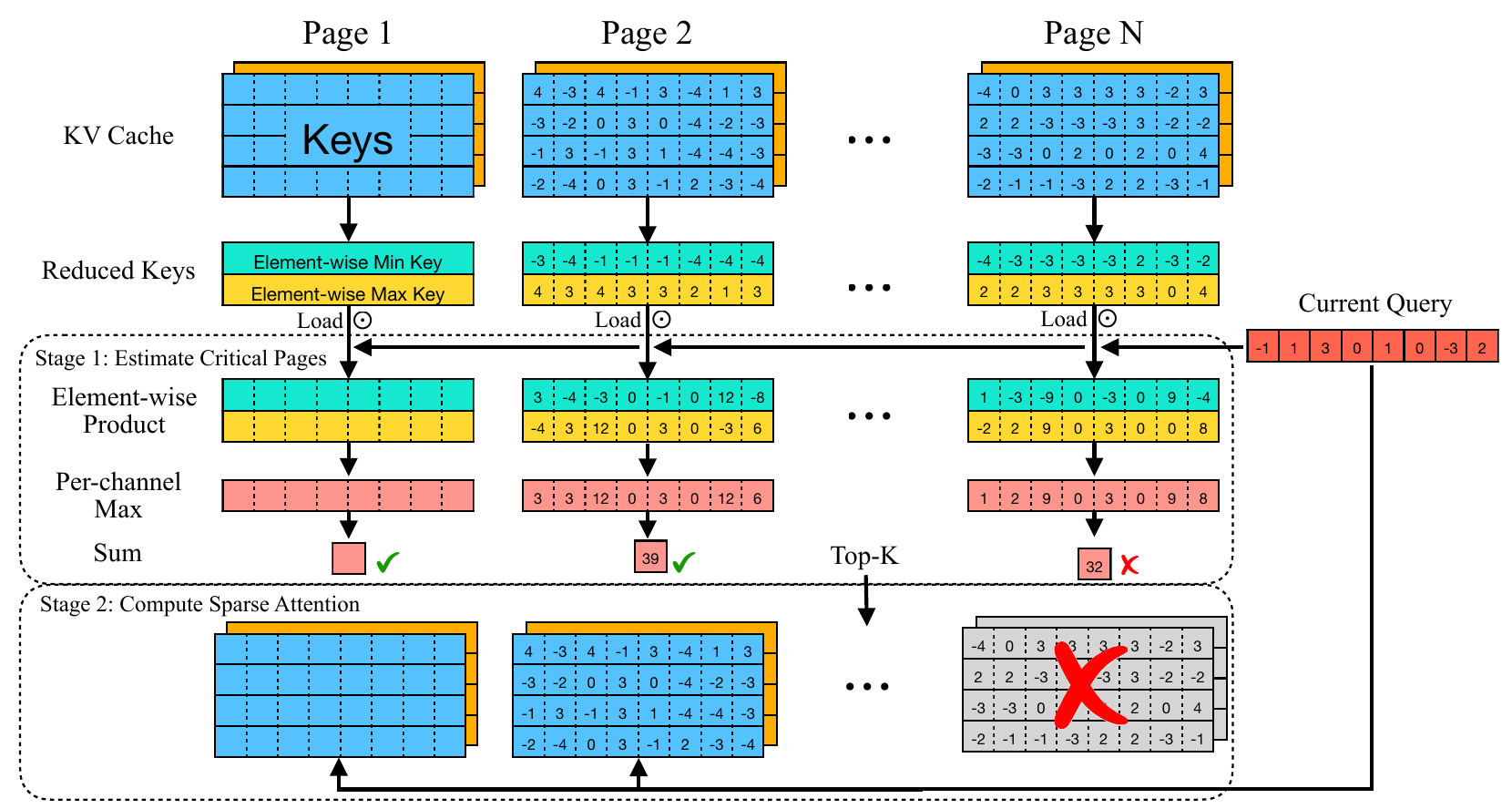}
    \caption{\method performs self-attention in two stages. In stage 1, \method estimates the criticality of pages by performing element-wise product between the current Query vector and both Min Key and Max Key vectors in each KV cache page. \method gets the sum of the per-channel maximal value for each page as the page criticality estimation. In stage 2, only Top-K KV cache pages are loaded to perform sparse self-attention with the current Query. }
    \label{fig:select}
\end{figure*}

\subsection{Self-Attention Operation Features High Sparsity}

Luckily, previous research has highlighted the inherent sparsity in self-attention~\cite{zhang2023h2o, ge2024model}. Due to this property of self-attention, a small portion of tokens in the KV cache, called critical tokens, can accumulate sufficient attention scores, capturing the most important inter-token relationships.
For example, as shown in \fig\ref{fig:qa-sparsity}, apart from the first two layers, less than 10\% of the tokens are needed to achieve similar accuracy, which makes the attention on the rest of the tokens unnecessary. 
Therefore, if we can estimate the criticality of the tokens, we can only compute self-attention on critical \kvc{} tokens to greatly reduce the memory movement and thus improve efficiency.

\subsection{Critical Tokens Depend on the Query}
However, the criticality of the tokens is dynamic and highly dependent on the query vector $Q$. Assuming the prompt is "A is B. C is D. A is", we demonstrate the attention map of a certain head in the 16th layer of Llama-2-7b in \fig~\ref{fig:query}. Since the output answer here should be "B", the token "B" is critical to the current query "is". Thus, it has a high attention score. However, before the final token "is", "B" is not critical for any previous query and has very low attention scores. In other words, the criticality of tokens is tightly related to the query token.

We quantify this effect by profiling the average recall rate of tokens with Top-10 attention scores along the text generations. The original attention with full \kvc{} can maintain $100$\% recall rate. However, \kvc{} eviction algorithm like H2O~\cite{zhang2023h2o} which prunes tokens based on history information, suffers from low recall rates since critical tokens are pruned in previous iterations. As shown in \fig\ref{fig:recall}, \method{} maintains recall rate close to full attention, as it estimated critical tokens based on current query. Therefore, pre-determining the criticality is challenging, which motivates query-aware sparsity by considering $Q$ vectors for criticality estimation.

\subsection{Dynamically Estimating Token Criticality}
\label{sec:methodDesp}
To efficiently and accurately estimate the criticality of \kvc{} tokens, we propose \method, an efficient and accurate algorithm that exploits query-aware context sparsity, which approximately selects the most potentially critical \kvc{} pages for the current query. We show the workflow of \method in \fig\ref{fig:select}. To manage the overhead, \method adopts PageAttention~\cite{kwon2023efficient} and selects the \kvc{} pages at the granularity of pages. 

To estimate the criticality of the pages, \method performs an approximate calculation of attention weights before the original attention operation, as shown in Algorithm \ref{algo:max}.

Our insight is that in order not to miss critical tokens, we should select pages containing the token with the highest attention weights. However, for an efficient selection of pages, we should calculate an approximate attention score following this insight. We found that the upper bound attention weights within a page can be used to approximate the highest attention in the page. The upper bound of the attention weights can be calculated by the channel-wise minimal values ($m_i$) and maximal values ($M_i$) of Key vectors. Given a $Q$ vector, \method calculates the maximum possible value of the channel $i$ by taking $U_i = \max(Q_i m_i, Q_i M_i)$. Note that $U_i$ is always greater than any product of $Q_i$ with the Key value $K_i$ for all tokens in this page regardless of the sign of $Q_i$. Therefore, when we add up $U_i$, we get the upper bound of attention weights across all Key vectors on this page.

After deriving the upper bound attention weights, we choose the top $K$ pages as critical, where $K$ is an arbitrarily defined hyper-parameter. To demonstrate the feasibility of \method, we perform actual self-attention and gather Top-K per-page attention scores. As shown in \fig\ref{fig:qa-sparsity}, our query-aware sparsity mostly aligns with the oracle sparsity. \method performs normal self-attention only on selected pages, which greatly reduces memory movement. We define the number of tokens in selected pages as the ``Token Budget''.

Due to the low sparsity ratio for the first two layers (as shown in \fig\ref{fig:qa-sparsity}), we only apply \method and all baselines on later layers to better preserve model accuracy. Note that whether to skip the first two layers or not is orthogonal to the \kvc{} selection algorithm. 
\begin{algorithm}[h]
   \caption{Token Criticality Estimation}
\begin{algorithmic}
    \label{algo:max}
   
   \STATE {\bfseries When inserting new token to KV cache:}
   \STATE {\bfseries Input:} Key vector $K$, Dimension of hidden states $dim$, Current maximal vector $M_i$, Current minimal vector $m_i$
   \STATE
   \FOR{$i=1$ {\bfseries to} $dim$}
        \STATE $M_i$ = $\max(M_i, k_{i})$     
        \STATE $m_i$ = $\min(m_i, k_{i})$     
    \ENDFOR
    \STATE
   \STATE {\bfseries When perform self-attention:}
   \STATE {\bfseries Input:} Query vector $Q$, Dimension of hidden states $dim$, Current maximal vector $M_i$, Current minimal vector $m_i$
   \STATE
   \STATE Initialize $score = 0$.
   \FOR{$i=1$ {\bfseries to} $dim$}
   \STATE $score$ += $MAX( q_i * max, q_i * min)$
   \ENDFOR
\end{algorithmic}
\end{algorithm}

\subsection{\method Reduces the Memory Movement of Self-Attention}

Instead of loading the whole \kvc{}, \method{} only needs to load a fraction of the data, which leverages \qaware{} sparsity. Assume that every $K$ or $V$ vector is $M$ bytes, the \kvc{} contains $L$ tokens, and each page contains $S$ KV pairs (Page size). During criticality estimation, \method will load maximal and minimal vectors of each page, which is approximately $2M*L/S$ bytes. Additionally, \method performs normal self-attention for top $K$ pages, which is $2M*K*S$ bytes. The whole \kvc{} is $2M*L$ bytes, which indicates \method loads $1/S + K*S/L$ of the total KV cache\footnote{The top-K operator incurs negligible memory loading and execution time (5-10 us). Therefore, we do not include it in efficiency analysis. }, which is equivalent to 
$$
\frac{1}{\text{Page Size}} + \frac{K}{\text{Page Num}}
$$

Assuming that we use $16$ KV pairs per page, context length is 64K, and we choose the top 4K pages, \method will reduce the memory load by $8\times$. Note that this memory load reduction is universal across all models and is compatible with existing quantization mechanisms~\cite{zhao2023atom}.

\section{Experiments}

\begin{table}
\small
    \setlength{\tabcolsep}{3pt}
    
    \centering
    \begin{tabular}{llccccc}
        \toprule
        \textbf{Method / Budget}  & 32 & 64 & 128 & 256 & 512 \\  \midrule
        H2O & 0\% & 1\% & 1\% & 1\% & 3\% \\
        TOVA & 0\% & 1\% & 1\% & 3\% & 8\% \\
        StreamingLLM & 1\% & 1\% & 1\% & 3\% & 5\% \\
        \textbf{\method (ours)} & \textbf{65\%} & \textbf{99\%} & \textbf{99\%} & \textbf{99\%} & \textbf{100\%} \\
        \bottomrule
    \vspace{0.03in}
    \end{tabular}
    \begin{tabular}{llccccc}
        \toprule
        \textbf{Method / Budget}  & 256 & 512 & 1024 & 2048 & 4096 \\  \midrule
        H2O & 2\% & 2\% & 2\% & 2\% & 4\% \\
        TOVA & 2\% & 2\% & 2\% & 2\% & 10\% \\
        StreamingLLM & 1\% & 1\% & 1\% & 2\% & 4\% \\
        \textbf{\method (ours)} & \textbf{88\%} & \textbf{92\%} & \textbf{96\%} & \textbf{100\%} & \textbf{100\%} \\
        \bottomrule
    \end{tabular}
    
    \caption{(i) Results of 10k length passkey retrieval test on LongChat-7b-v1.5-32k. (ii) Results of 100k length passkey retrieval test on Yarn-Llama-2-7b-128k. Quest can achieve nearly perfect accuracy with 64 and 1024 tokens KV cache budget, which is about 1\% of the total sequence length, demonstrating that \method can effectively preserve the model's ability to handle long-dependency tasks. However, KV cache eviction algorithms such as H2O, TOVA, and StreamingLLM incorrectly discard the KV cache of the answer before receiving the question, thus failing to achieve ideal accuracy. }
    \label{tab:passkey}
\end{table}

\subsection{Setting}
We evaluate \method on the language modeling dataset PG19~\cite{raecompressive2019}, passkey retrieval task~\cite{peng2023yarn}, and six datasets in LongBench~\cite{bai2023longbench}: NarrativeQA~\cite{kocisky-etal-2018-narrativeqa}, HotpotQA~\cite{yang2018hotpotqa}, Qasper~\cite{dasigi2021dataset}, TrivialQA~\cite{joshi-etal-2017-triviaqa}, GovReport~\cite{huang-etal-2021-efficient}, MultifieldQA~\cite{bai2023longbench}. We choose two widely used long-context models for our evaluation: LongChat-v1.5-7b-32k~\cite{longchat2023} and Yarn-Llama-2-7b-128k~\cite{peng2023yarn}. We compare our method against the KV cache eviction algorithm H2O~\cite{zhang2023h2o}, TOVA~\cite{oren2024transformers}, and StreamingLLM~\cite{xiao2023streamingllm}. Note that we \textbf{do not} apply any \method and other baseline algorithms to the first two layers of the model, as our analysis in Sec~\ref{sec:methodDesp} indicates a low sparsity ratio for these layers.

\subsection{Accuracy Evaluation}
\subsubsection{Language Modeling on PG19}

\begin{figure}
    \centering
     \includegraphics[width=0.45\textwidth]{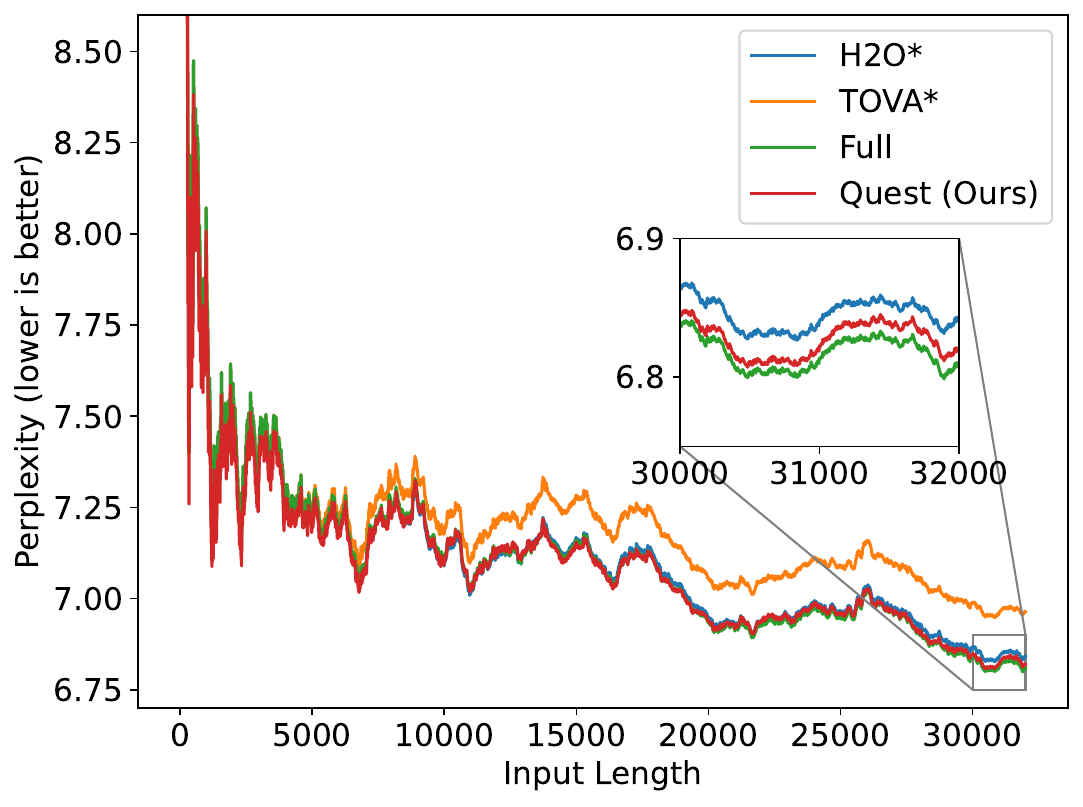}
    \caption{Language modeling evaluation of \method on PG19 dataset. We prompt the model with 0 to 32000 tokens from the PG19 test set and measure the perplexity of output tokens. H2O* and TOVA* indicate that for the first two layers of models, we do not apply these two algorithms to prune the KV Cache, as analyzed in Sec~\ref{sec:methodDesp}, which better preserves the model performance. \method also uses a full cache in the first two layers of the model. \method can closely match the performance of the full cache model.}
    \label{fig:pg19ppl}

\end{figure}

We first evaluate the language modeling perplexity on the PG19 test set, which is a dataset comprising 100 books with an average length of 70k tokens. We use the LongChat-7b-v1.5-32k model to test 32k tokens on PG19. We feed the model with various numbers of tokens and evaluate the perplexity of generated tokens. We evaluate H2O, TOVA, and \method with a token budget of 4096, which is approximately 1/8 of the total token length.  As indicated by the perplexity results in \fig\ref{fig:pg19ppl}, \method's accuracy closely matches the oracle baseline with a full KV cache.

\subsubsection{Results on long text passkey retrieval task}
Since language modeling evaluation only involves local dependencies, models can achieve great performance by focusing on recent tokens. However, the ability to handle long-distance dependencies is crucial for long text reasoning. For KV cache eviction algorithms like H2O and TOVA, parts of KV caches that are important for distant future tokens may be discarded, thereby preventing the model from obtaining the correct answer.

To show that \method helps maintain the ability of models to handle longer dependency tasks, we evaluate it on the passkey retrieval task from Yarn~\cite{peng2023yarn}. This task measures a model's ability to retrieve a simple passkey from a large amount of meaningless text. We put the answer in different depth ratios of the text and evaluate if the model can retrieve the correct answer with different KV cache token budgets. We evaluate LongChat-7b-v1.5-32k on 10k tokens test and Yarn-Llama-2-7b-128k on 100k tokens test.

Since H2O~\cite{zhang2023h2o} needs to calculate historical attention scores for KV cache pruning, it needs to compute the complete $O(n^2)$ attention map and thus is unable to use Flash-Attention~\cite{dao2022flashattention} for long-context inference. Therefore, to enable H2O on long-context evaluation, we use Flash-Attention in the context stage for the 100k sequence length passkey retrieval test and start collecting historical attention scores for H2O in the decoding stage. For TOVA~\cite{oren2024transformers} and StreamingLLM~\cite{xiao2023streamingllm}, we evaluated them on the 10k and 100k sequence lengths.

\begin{figure*}[t]
    \centering
    \subfigure{
        \includegraphics[width=0.31\linewidth]{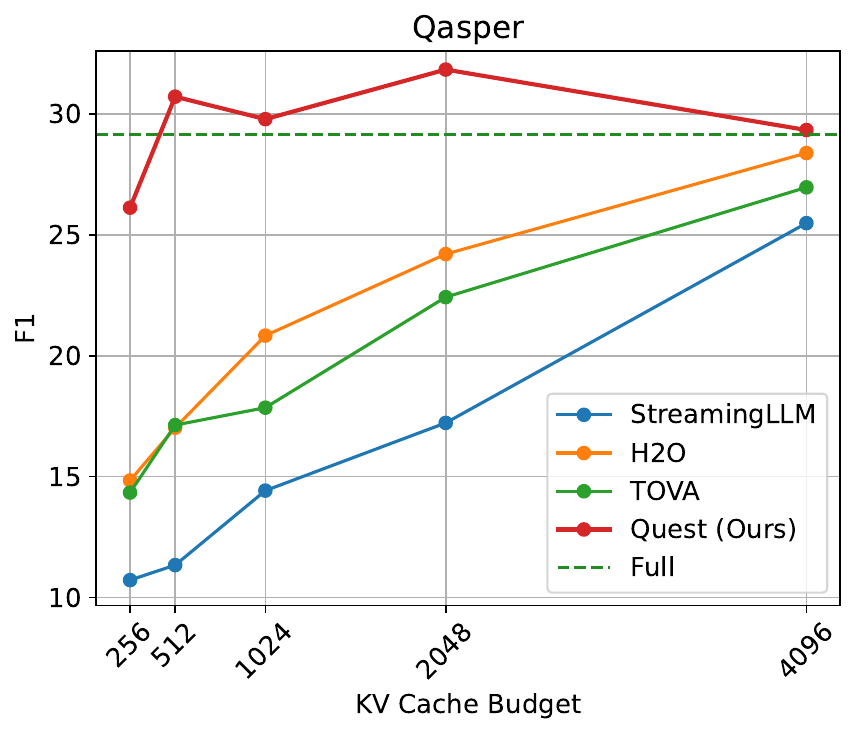}
    }
    \subfigure{
        \includegraphics[width=0.31\linewidth]{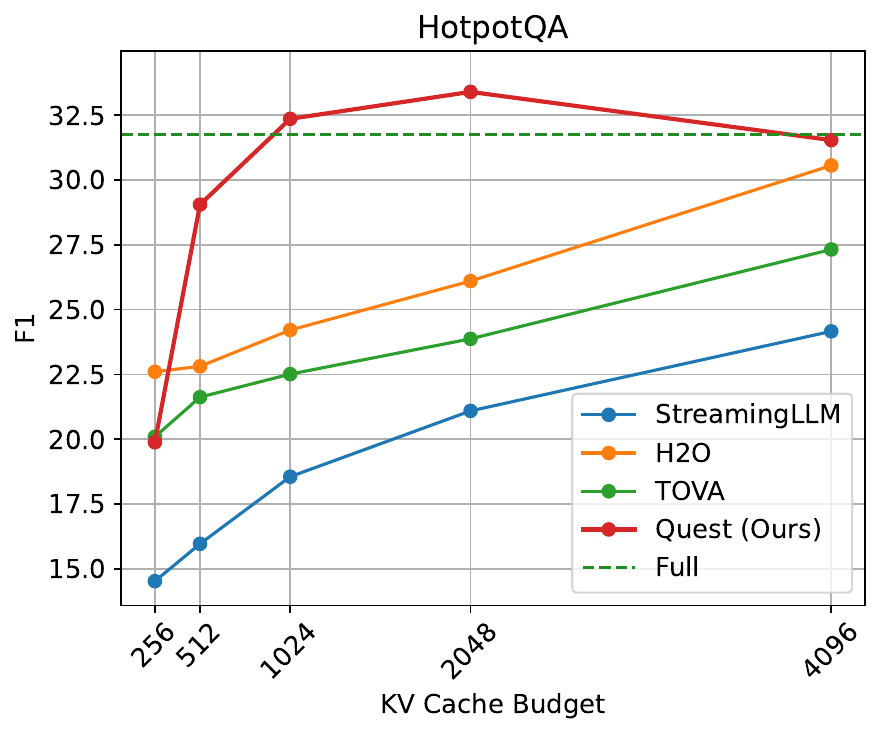}
    }
    \subfigure{
        \includegraphics[width=0.31\linewidth]{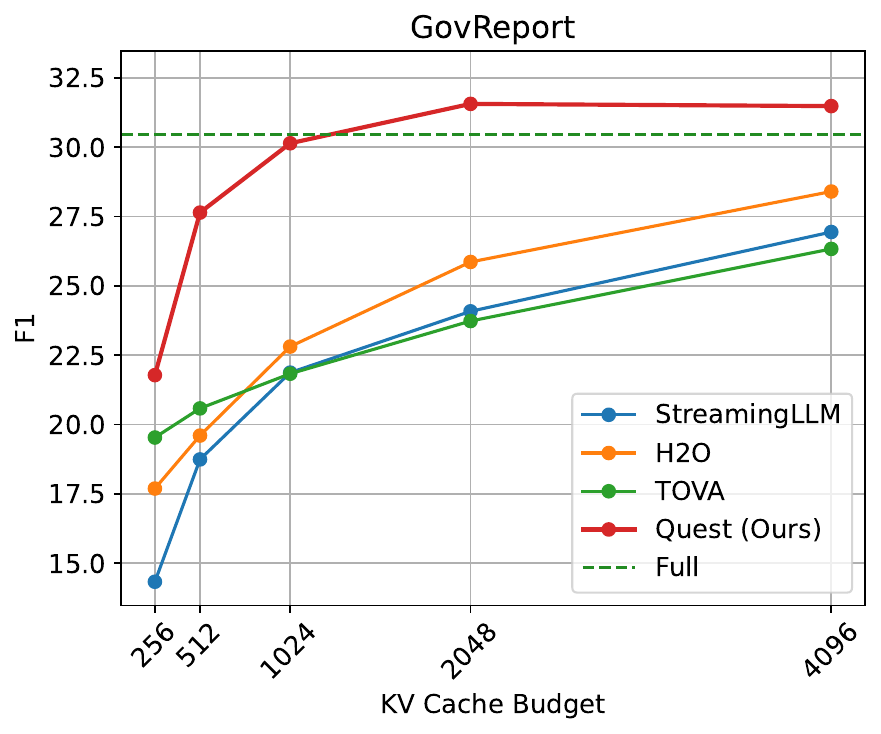}
    }
    \subfigure{
        \includegraphics[width=0.31\linewidth]{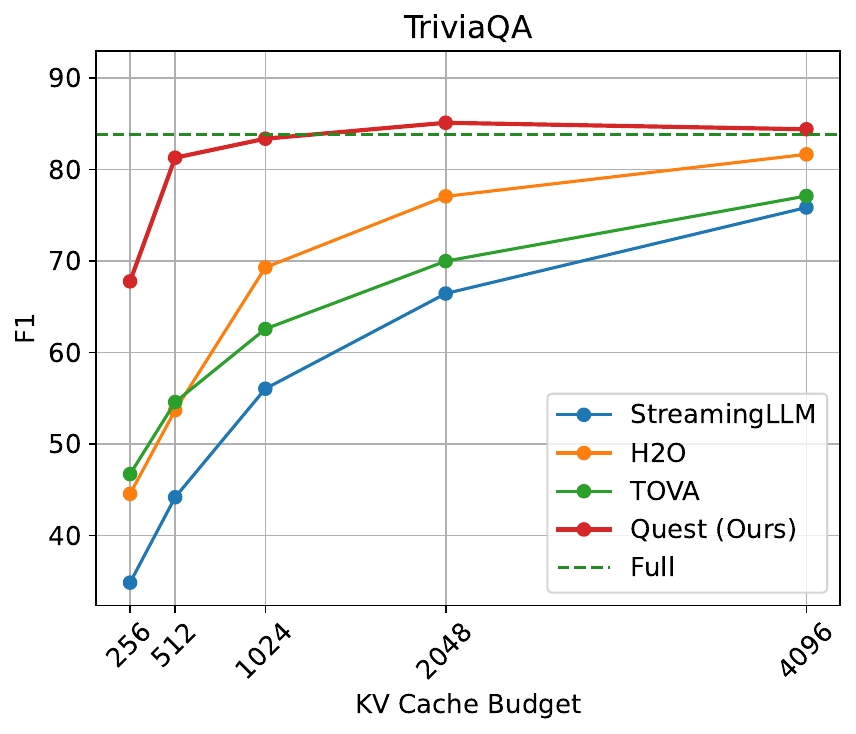}
    }
    \subfigure{
        \includegraphics[width=0.31\linewidth]{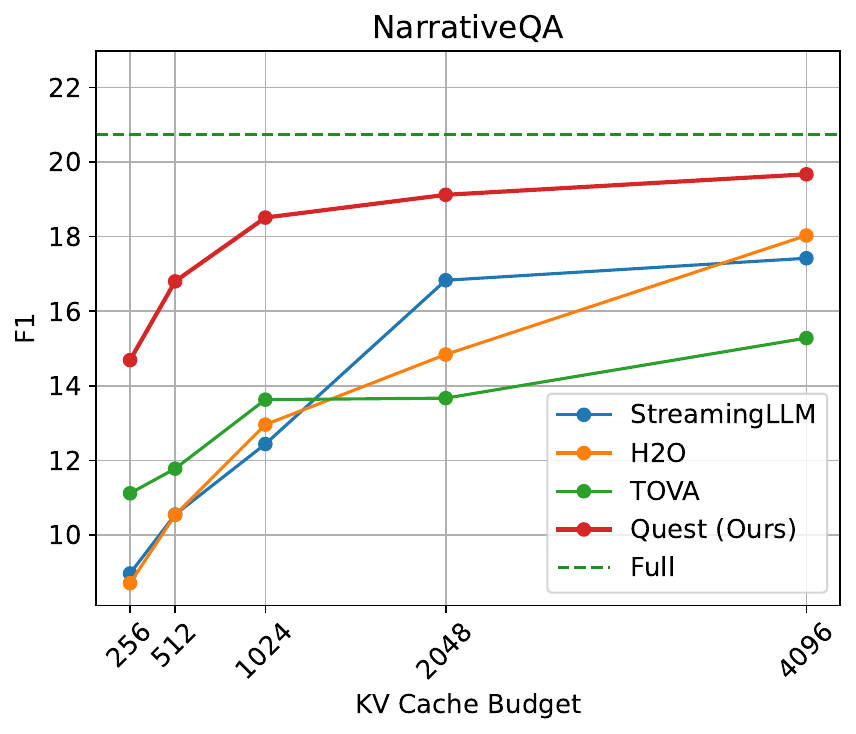}
    }
    \subfigure{
        \includegraphics[width=0.31\linewidth]{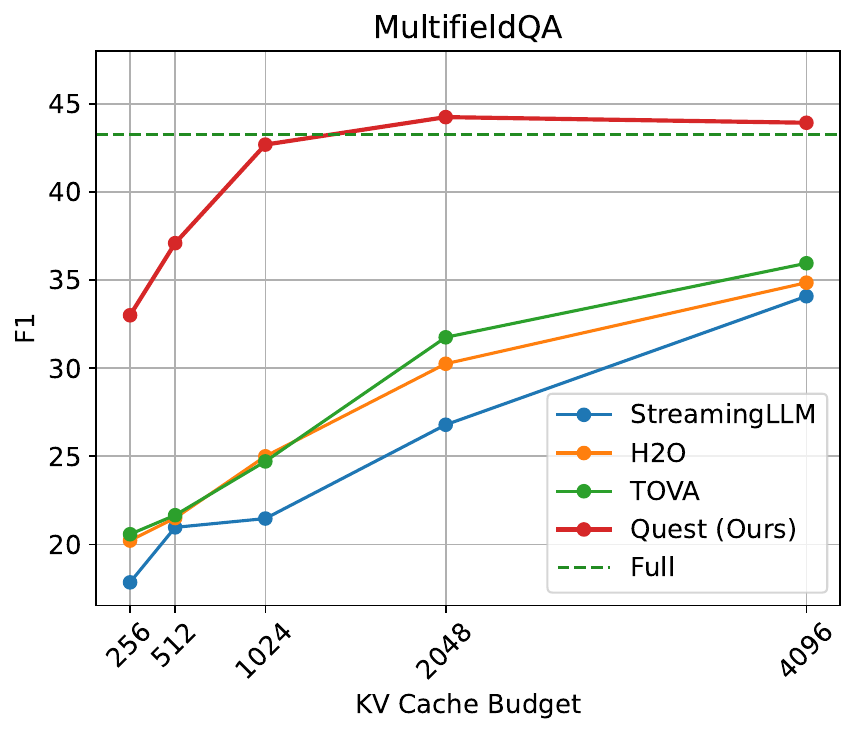}
    }
    \caption{We evaluate \method and baselines across six long context datasets with various token budgets. \method constantly surpassing all baselines at all datasets and all token budgets. For most of the dataset, \method reaches comparable accuracy with a $1$K token budget. To evaluate the impact of different methods on the model's ability to retrieve long-dependency information, we simulate decoding by feeding the task's question to the model token by token. }
    \label{fig:longbench}
\end{figure*}

For the passkey retrieval test, we directly prefill the input text containing the passkey and texts to the model. However, to evaluate the impact of different methods on the model's ability to handle long-dependency tasks in practical scenarios, we simulate decoding by feeding the task's question and instruction to the model token by token. In this case, H2O and TOVA might mistakenly discard tokens critical for future tokens, such as the passkey that will be queried later. Similarly, StreamingLLM can only focus on the most recent text window, and if the passkey appears outside this window, it cannot provide the correct answer. Therefore, H2O, TOVA, and StreamingLLM cannot achieve ideal accuracy on the 10k and 100k length passkey retrieve test. However, \method does not discard KV cache but instead uses a query-aware approach to identify critical tokens. As shown in Tab.~\ref{tab:passkey}, \method can achieve perfect accuracy with a minimal budget both on 10k and 100k sequence length tests.

\begin{figure*}[t]
    \centering
    \subfigure[\estimate{}]{
        \includegraphics[width=0.47\linewidth]{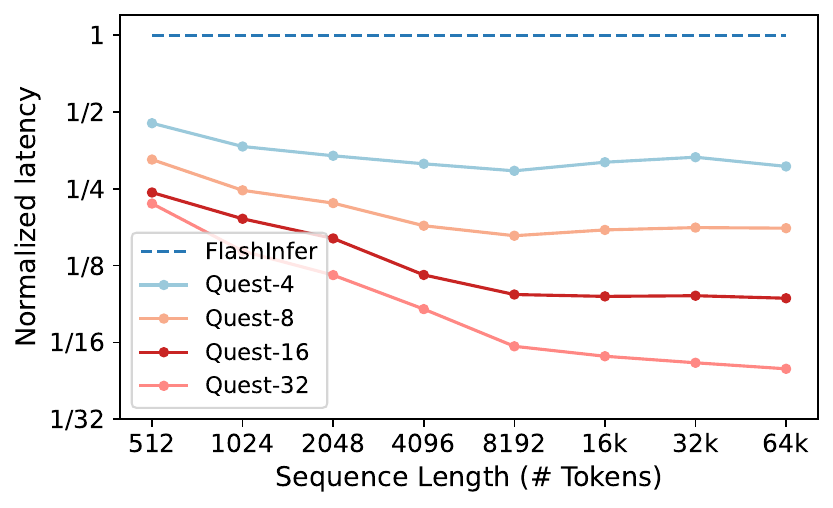}
        \label{fig:kernel-estimate}
    }
    \subfigure[\approxattn{}]{
        \includegraphics[width=0.47\linewidth]{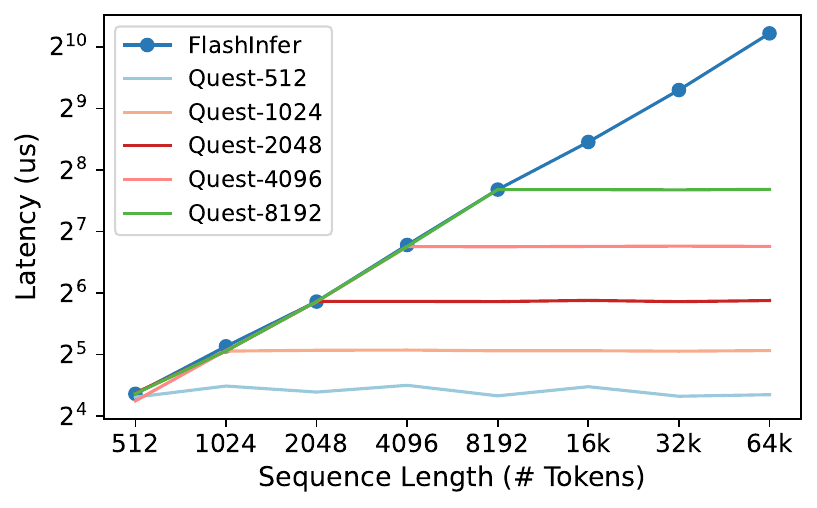}
        \label{fig:kernel-approxattn}
    }
    \caption{We measure the latency of individual kernels in \method. (a) As sequence length increases, the relative criticality estimation latency decreases to $1/\text{Page Size}$ of FlashInfer. (b) Approximate attention with token budget K consumes constant time irrelevant to total sequence length and reaches similar performance of FlashInfer at sequence length K.}
    \label{fig:kernel_efficiency}
\end{figure*}

\subsubsection{Results on LongBench}

To validate that \method can outperform baselines on general long-context datasets, we evaluate our method and baselines on six datasets in LongBench. We evaluate on LongChat-7b-v1.5-32k across a wide range of long-context datasets, including single-document QA: NarrativeQA, Qasper, MultiFieldQA; multi-document QA: HotpotQA; summarization: GovReport; few-shot learning: TriviaQA. We evaluate H2O, TOVA, StreamingLLM, and \method with different KV cache budgets. For all datasets, we split the input into material and question/instruction. For the material part, we use Flash-Attention~\cite{dao2022flashattention} with the full KV cache to perform inference. For the question part, we simulate decoding by feeding them to the model token by token. Similar to the passkey retrieval test, to enable H2O to use Flash-Attention, we could not collect H2O's historical attention scores during the context stage, thus starting from the decoding stage.

As shown in the \fig\ref{fig:longbench}, \method consistently outperforms all baselines across six long-context datasets with various KV cache budgets. \method with a budget of $1$K tokens can achieve comparable performance as the model with full KV cache, while other baselines still exhibit a notable gap from full cache performance even with a larger budget. After considering the full cache used in the first two layers, \method can achieve lossless performance on Qasper, HotpotQA, GovReport, TriviaQA, NarrativeQA, and MultifieldQA with KV cache sparsity of 1/6, 1/6, 1/5, 1/10, 1/5, and 1/6, respectively. This demonstrates that \method is capable of maintaining the model's capabilities across different types of long-context tasks, as it does not lead to the generation of incorrect answers due to improper discarding of KV cache.

\subsection{Efficiency evaluation}
To demonstrate the feasibility of \method{}, we implement the entire framework with dedicated CUDA kernels based on FlashInfer~\cite{flashinfer}, a kernel library for LLM inference. We first evaluate \method{}'s kernel-level efficiency under the configuration of Llama2-7B on an RTX4090 with CUDA 12.2 in Sec~\ref{sec:eval:kernel}. Besides, we show the end-to-end speedup of \method{} in text generation as shown in Sec~\ref{sec:eval:e2e}. We compare \method{} with a normal attention implementation from the original FlashInfer. To demonstrate the improvement, we qualitatively compare efficiency under the same accuracy between \method{} and baselines in Sec~\ref{sec:eval:comparison}. Note that we use an Ada 6000 GPU~\cite{ada6000} in end-to-end evaluations for longer context length.

\subsubsection{Kernel evaluation}
\label{sec:eval:kernel}
Due to the memory-bound nature of LLM inference, the speedup of \method{} is proportional to the sparsity ratio (which is equivalent to memory movement reduction). We quantify this effect in \fig\ref{fig:kernel_efficiency}, which evaluates per-kernel performance with NVIDIA's benchmark tool NVBench~\cite{nvidia_nvbench}.

\textbf{\estimate{}}
We evaluate the latency of criticality estimation in \method{} under different sequence lengths and page sizes. At short sequence length, the memory bandwidth utilization of estimation is smaller than that of FlashInfer, as the total memory load size is not enough to fully utilize GPU memory bandwidth. As sequence length grows, the relative performance improves and approaches $1/\text{Page Size}$ since estimation only consumes one token per page. Note that techniques like quantization or larger page size can further reduce the additional memory usage.

\begin{figure*} [t]
    \centering
     \includegraphics[width=\textwidth]{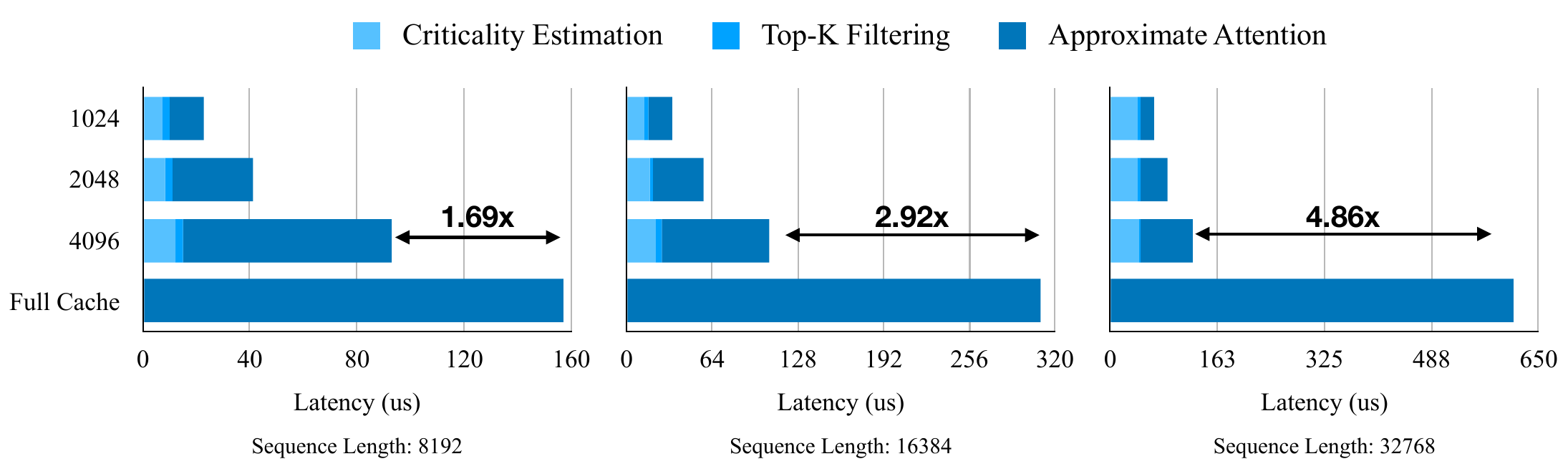}
    \caption{\method self-attention time breakdown compared to FlashInfer. At all sequence lengths, \method significantly outperforms FlashInfer, as the memory movement is reduced. At sequence length $32$K with token budget $2048$, \method speeds up self-attention by \selfspeedup.}
    \label{fig:effibreak}
\end{figure*}

\begin{figure*}
    \centering
     \includegraphics[width=0.97\textwidth]{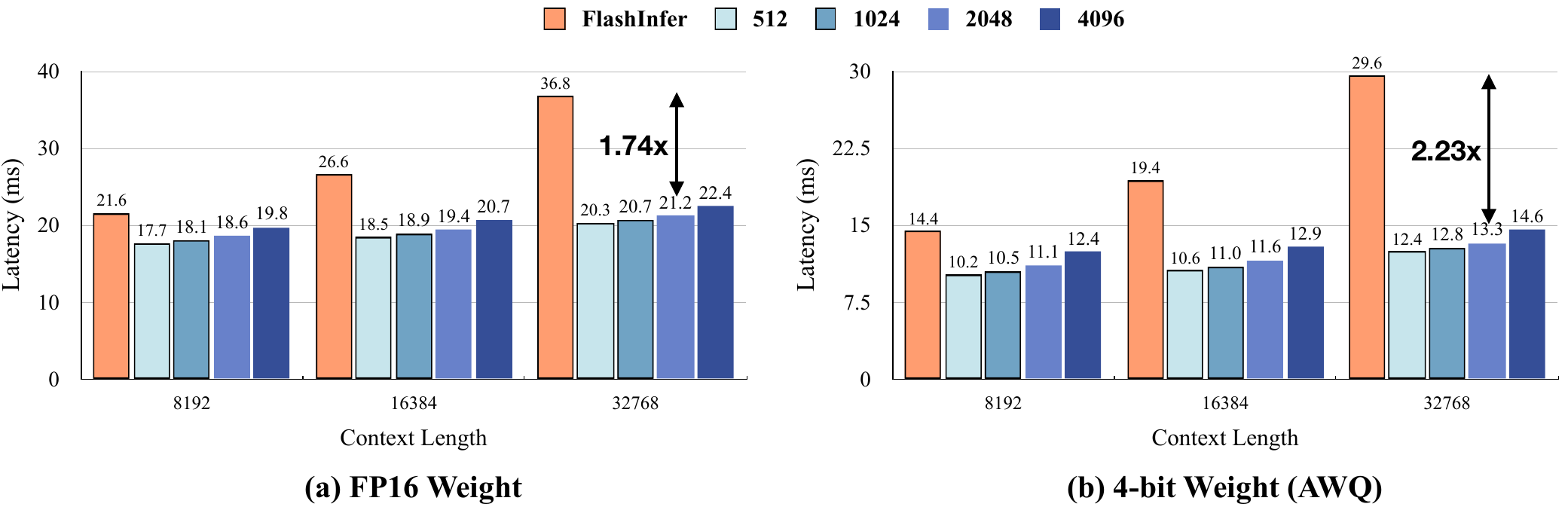}
    \caption{End-to-end latency of \method. For all sequence lengths, \method significantly outperforms FlashInfer. Increasing the sequence lengths only slightly changes the latency of \method. At a given sequence length, \method's latency slightly increases as the token budget grows. With sequence length $32$K, token budget $2048$, $4$-bit weight quantization, \method speedup end-to-end inference by \eespeedup. }
    \label{fig:e2e}
\end{figure*}

\textbf{\topk{}}
We enable the \topk{} in \method{} with a batched Top-K CUDA operator from a vector search kernel library RAFT~\cite{topk2023}. We test the latency of \topk{} under different sequence lengths and token budgets. Since \estimate{} reduces one entire token into one criticality score, \topk{} has limited memory movement compared to other operators, thus having a low latency overhead of 5-10 us for sequence length less than $128$k.

\textbf{\approxattn{}}
Since \method{} is compatible with PageAttention, approximate attention can be easily implemented by feeding Top-K page indices as sparse loading indices. We compare \method{}'s approximate attention with the original attention of FlashInfer under different sequence lengths and token budgets with a $16$ page size. At a given token budget $B$, the latency of \approxattn{} is a constant regardless of the sequence length. Since \approxattn{} introduces minimal overhead, it has a similar latency as FlashInfer at sequence length $B$.

We further evaluate \method{}'s attention mechanism, which combines \estimate{},\topk{}, and \approxattn{}, on the Llama2-7B model using the PyTorch profiler. We show the time breakdown of \method{} in \fig\ref{fig:effibreak} on various sequence lengths. \method{} reduce the self-attention time by \selfspeedup compared with FlashInfer at $32$K sequence length with $2048$ token budget.

\begin{figure*}[!h]
    \centering
     \includegraphics[width=0.99\textwidth]{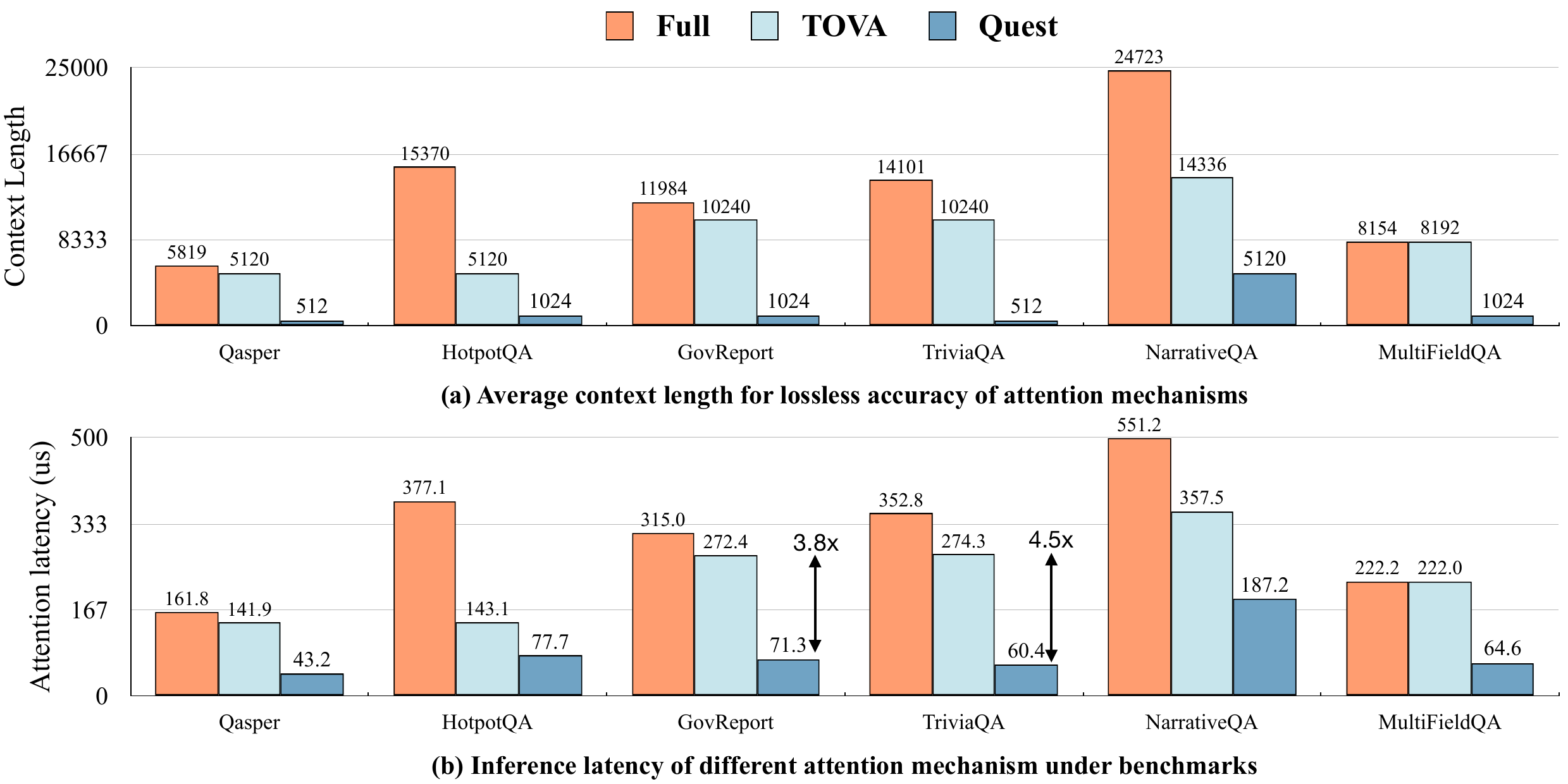}
    \caption{Efficiency comparison of \method{} with baselines under the same accuracy constraint. (a) Tokens budgets needed for comparable accuracy by different attention methods. Full denotes the original attention, which means the average context length of benchmarks. (b) Inference latency of different attention methods for comparable accuracy. \method{} boosts $3.82\times$ speed on GovReport compared to TOVA.}
    \label{fig:comparision}

\end{figure*}

\subsubsection{End-to-End Evaluation}
\label{sec:eval:e2e}
To show the practical speedup of \method{}, we deploy the framework into real-world single-batch scenarios. We measure the average latency of generating one token in the decode stage under different sequence lengths and token budgets. Note that we do not measure the sampling process since its execution time is smaller and depends on the setting. We compare \method{} with a full \kvc{} baseline which is implemented by FlashInfer. As shown in \fig\ref{fig:e2e}, \method{} outperforms FlashInfer at all sequence lengths. The latency of \method{} grows significantly slower than FlashInfer when the sequence length increases, as \method{} maintains similar token budgets. At sequence length $32$K and token budget $2048$, \method{} boosts inference speed by $1.74\times$ with FP16 weights and $2.23\times$ with 4-bit quantized weight.

\subsubsection{Comparison with Baselines}
\label{sec:eval:comparison}
To demonstrate the performance improvements of \method{}, we compare the inference efficiency of different attention mechanisms under the same accuracy constraint, i.e. lossless accuracy of six tasks from LongBench. We show token budgets needed for the lossless accuracy target by different attention mechanisms in Fig~\ref{fig:comparision}(a). For example, NarrativeQA exhibits an average context length of $24$K tokens. To achieve lossless accuracy, TOVA requires a token budget of $14$K, whereas \method{} necessitates only $5$K tokens leading to much higher sparsity.

However, none of the baselines included a kernel implementation of their proposed method. Consequently, we conduct a qualitative analysis of the baselines' self-attention efficiency by utilizing the inference latency of FlashInfer, disregarding other runtime overheads (e.g., TOVA's requirement to calculate history scores~\cite{oren2024transformers}). In contrast, \method{} is evaluated in a practical setting with consideration of all operators. As shown in \fig\ref{fig:comparision}(b), \method{} significantly surpasses all baselines in terms of self-attention latency due to the high query-aware sparsity. For GovReport and TriviaQA, \method{} boosts the inference by $3.82\times$ and $4.54\times$, respectively. Therefore, \method{} can achieve higher efficiency while maintaining superior accuracy.

\section{Conclusion}
We present \method, an efficient and accurate \kvc{} selection algorithm that exploits \qaware{} sparsity. \method{} dynamically estimates the criticality of tokens in \kvc{} based on the per-page metadata and the current query. It then performs self-attention only on the critical tokens with greatly reduced memory movement, providing high sparsity with negligible accuracy loss. Comprehensive evaluations demonstrate that \method provides up to \selfspeedup self-attention speedup, which contributes to \eespeedup end-to-end latency reduction in the decode phase. Compared to prior baselines, \method{} reduces up to $4.5\times$ self-attention latency with the same accuracy target under long-context benchmarks.

\section*{Acknowledgements}
We thank MIT-IBM Watson AI Lab, Amazon and MIT Science Hub, MIT AI Hardware Program, National Science Foundation, and Samsung for supporting this research. We thank NVIDIA for donating the DGX server. 
We thank Zihao Ye for his insightful discussion, feedback, and useful advice on algorithm design and FlashInfer integration. This work was also supported by
generous gifts from Intel, Google, and the PRISM Research Center, a JUMP Center cosponsored by SRC and DARPA.

\bibliography{main}
\bibliographystyle{icml2024}

\end{document}